\documentclass[lettersize,journal]{IEEEtran}
\usepackage{amsmath,amsfonts}
\usepackage{algorithmic}
\usepackage{algorithm}
\usepackage{array}
\usepackage[caption=false,font=normalsize,labelfont=sf,textfont=sf]{subfig}
\usepackage{textcomp}
\usepackage{stfloats}
\usepackage{url}
\usepackage{verbatim}
\usepackage{graphicx}
\usepackage{cite}
\usepackage{booktabs}
\usepackage[table,xcdraw]{xcolor}
\usepackage{multirow}
\usepackage{tikz}
\begin{document}

\title{From Coarse to Nuanced:  Cross-Modal Alignment of Fine-Grained Linguistic Cues and Visual Salient Regions for Dynamic Emotion Recognition}

\author{Yu~Liu,
        Leyuan~Qu,
        Hanlei~Shi,
        Di~Gao,
        Yuhua~Zheng,
        Taihao~Li
\thanks{This work was supported in part by the Scientific Research Staring Foundation of Hangzhou Institute for Advanced Study (2024HIASC2001), in part by Zhejiang Provincial Natural Science Foundation of China (No. LQN25F020001), and in part by the Key R\&D Program of Zhejiang (2025C01104). \textit{(Corresponding author: Taihao~Li, Leyuan~Qu)}}
\thanks{All authors are with Hangzhou Institute for Advanced Study, University of Chinese Academy of Sciences, Hangzhou, China.}
\thanks{Yu~Liu (e-mail: liuyu233@mails.ucas.ac.cn), Leyuan~Qu (e-mail: leyuan.qu@ucas.ac.cn), Hanlei~Shi (e-mail: shihanlei23@mails.ucas.ac.cn), Di~Gao (e-mail: gaodi@ucas.ac.cn), Yuhua~Zheng (e-mail: zhengyuhua@ucas.ac.cn), and Taihao~Li (e-mail: lith@ucas.ac.cn).}

}

\maketitle

\graphicspath{{./figs/}}

\begin{center}
  \textit{This work has been submitted to the IEEE for possible publication. Copyright may be transferred without notice, after which this version may no longer be accessible.}
\end{center}

\begin{abstract}
Dynamic Facial Expression Recognition (DFER) aims to identify human emotions from temporally evolving facial movements and plays a critical role in affective computing. While recent vision-language approaches have introduced semantic textual descriptions to guide expression recognition, existing methods still face two key limitations: they often underutilize the subtle emotional cues embedded in generated text, and they have yet to incorporate sufficiently effective mechanisms for filtering out facial dynamics that are irrelevant to emotional expression. 
To address these gaps, We propose GRACE, \textbf{\underline{G}ranular \underline{R}epresentation \underline{A}lignment for \underline{C}ross-modal \underline{E}motion recognition} that integrates dynamic motion modeling, semantic text refinement, and token-level cross-modal alignment to facilitate the precise localization of emotionally salient spatiotemporal features. Our method constructs emotion-aware textual descriptions via a Coarse-to-fine Affective Text Enhancement (CATE) module and highlights expression-relevant facial motion through a motion-difference weighting mechanism. These refined semantic and visual signals are aligned at the token level using entropy-regularized optimal transport. Experiments on three benchmark datasets demonstrate that our method significantly improves recognition performance, particularly in challenging settings with ambiguous or imbalanced emotion classes, establishing new state-of-the-art (SOTA) results in terms of both UAR and WAR.
\end{abstract}


\section{Introduction}
\label{sec:intro}

\IEEEPARstart{D}{ynamic} facial expression recognition (DFER) focuses on inferring emotion from facial motion. It has evolved from controlled lab scenarios~\cite{lucey2010extended,valstar2010induced,zhao2011facial} to in-the-wild settings~\cite{jiang2020dfew, liu2022mafw, wang2022ferv39k} and supports diverse applications such as human-computer interaction and mental-health assessment~\cite{189ye2024dep,197jiang2023efficient,10zhao2016predicting}.
Early attempts in DFER follow a vision-only pipeline that progressively evolves from Convolutional Neural Networks (CNN)\cite{Yu2018,Liu2020SAANet}, to 3D CNN/Long Short-term Memory Networks (LSTM) hybrids\cite{3D-ResNet-18,former} that capture short-term motion, and finally to Transformer-style backbones\cite{former, Liu2023EST, Ma2023LOGOFormer} for long-range modeling. 
While these visual routines outperform static-image models by capturing spatiotemporal cues, they still rely on coarse one-hot labels-falling to encode \emph{what} muscular actions occur, \emph{how strongly}, and \emph{when}~\cite{Freq-HD, savchenko2023facial}. 
A more fundamental challenge is the presence of \textit{emotion-irrelevant motion}-where treating a snippet of video as a monolithic unit causes the model to latch onto spurious cues like blinks or head turns in some specific emotion~\cite{ben2021video}. 
Although enhancements such as optical flow and frequency-aware features have been introduced~\cite{Freq-HD,Poux2022OpticalFlow,Zhao2023STGraph,Liu2020SAANet,S2D}, performance on subtle or low-intensity expressions remains poor, suggesting the core issue persists.

\begin{figure}[!t]
\label{fig:overview_arch}
\centering
\includegraphics[width=\linewidth, trim=95 0 95 0, clip]{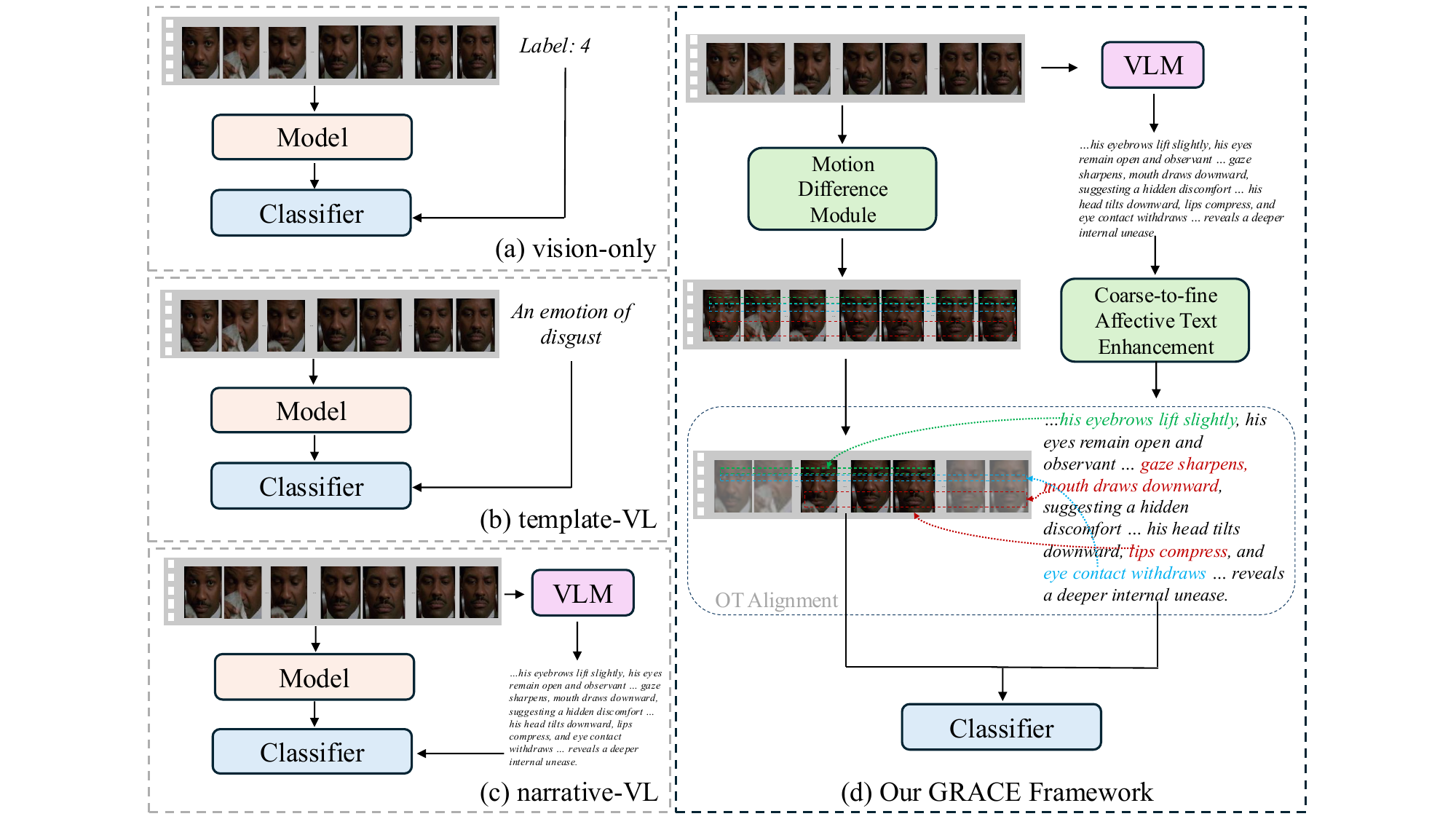}
\caption{Comparison of existing methods and our proposed GRACE. (a) Vision-only models directly process the video sequence and predict from a clip-level representation. (b) Template-based VL models introduce generic textual prompts but provide no detailed semantic grounding. (c) Narrative-VL models utilize descriptive captions, but typically compress the entire text into a global embedding and align it with pooled video features, ignoring fine-grained correspondence. (d) Our GRACE introduces motion-difference based filtering and token-level optimal transport (OT) alignment between text tokens refined by CATE and spatio-temporal video segments, enabling interpretable, emotion-sensitive prediction.}
\end{figure}

These limitations have sparked growing interest in vision–language (VL) models, which incorporate either predefined prompts or free-form descriptions to provide semantic guidance~\cite{wang2024survey}. 
By aligning facial motion with textual semantics, they promise a more fine-grained emotional perception. Early \emph{template-VL} models use fixed prompts such as
“an emotion of \emph{label}”~\cite{DFER-CLIP, emoclip, DK-CLIP, zhao2025enhancing, tao20243} to enable zero-shot inference, but they lack fine-grained action semantics. To overcome this limitation,
\emph{narrative-VL} models emerged~\cite{FineCLIPER}, which pairs clips with free-form descriptions. These descriptions explicitly specify \emph{what} muscular actions occur and \emph{how strongly}, offering richer, continuous supervision compared to coarse one-hot labels. 
Recent efforts have further explored how linguistic prompts can help disambiguate subtle expressions~\cite{13,140} or enable emotion-grounded alignment via textual anchors~\cite{FineCLIPER}. 

Despite theoretically richer supervision, these models yield only marginal empirical improvements. Notably, emotion-irrelevant motions, such as subtle facial movements or posture shifts not explicitly described, continue to persist and bias predictions. Consequently, the problem of \emph{emotion-irrelevant motion} remains unresolved, suggesting deeper structural limitations within current VL methods.

We argue that these persistent issues stem from two \emph{intertwined} limitations that reflect a deeper design bottleneck: First, they flatten each caption into a single embedding before alignment, collapsing the fine-grained semantic cues embedded within the sentence—an phenomenon we refer to as \emph{text-granularity loss}. Second, clip-level pooling on the video side merges all frames—expressive and neutral alike—into a single representation, allowing irrelevant motions like blinks or head turns to persist as false emotion cues. We term this residual noise \textit{emotion-irrelevant motion}.
At their core, both problems arise from a shared \textit{flatten-then-align} paradigm, which erases internal structure on both the language and visual sides. This discards critical cues—\textit{what} textual elements describe and \textit{where} visual changes occur—that are essential for precise semantic alignment. This raises a central question: can a granularity-preserving text representation, paired with a motion-aware video weighting scheme, enable finer-grained cross-modal alignment and mitigate both sources of noise?

We posit that overcoming the \textit{flatten-then-align} bottleneck requires preserving and leveraging multi-granular semantic structure—both in language and in spatio-temporal facial dynamics. Crucially, the effectiveness of maintaining granularity-aware semantics is not merely theoretical; it has been validated in other domains. In Natural Language Processing, granularity-aware encoders that keep sentence-, phrase-, and word-level cues have advanced tasks like discourse parsing and relation extraction~\cite{bert,liu2019hierarchical,hdt2024}. Similarly, in vision–language tasks such as visual question answering, preserving linguistic granularity helps guide attention to semantically relevant regions~\cite{nguyen2019mtl,li2020hero,ye2022hmn}. These findings converge on a key insight: preserving internal granularity enables dual capabilities—it clarifies \emph{what} to seek (semantic cues) and indicates \emph{where} to focus (salient frames or segments). 
Motivated by this evidence, we hypothesize that a granularity-preserving design can not only help DFER identify brief, localized facial expressions (“\textit{what}”), but also provide semantic anchors (“\textit{where}”) to isolate meaningful motion from \textit{emotion-irrelevant} noise. To this end, we propose a model that (i) retains multi-granular semantic cues within the textual descriptions and (ii) leverages these cues to selectively attend to emotion-bearing frames—precisely what we pursue next.

We therefore introduce \textbf{\underline{G}ranular \underline{R}epresentation \underline{A}lignment for \underline{C}ross-modal \underline{E}motion recognition (\textsc{GRACE})}, a cross-modal framework that integrates semantic granularity and motion dynamics through emotion-tuned textual guidance and token-level alignment.
It begins with a granularity-aware text conditioning module that refines the original caption using emotion-informed prompts, enhancing emotion-relevant expressions within the text. The resulting token sequence serves as a semantically enriched input, providing targeted lexical cues for alignment with spatiotemporal facial dynamics.
This process selectively amplifies action-related descriptions, enabling token-level semantic guidance that is emotionally discriminative.
To exploit this multi-granular representation, we formulate cross-modal alignment as an Optimal Transport (OT) problem, matching emotion-relevant textual cues (e.g., “raised eyebrows”) to spatio-temporal video segments by minimizing semantic transport cost.
Segments with no low-cost matches are naturally down-weighted—thus filtering out \textit{emotion-irrelevant motion} lacking textual grounding.
These sparse, high-confidence alignments form the basis of our final emotion prediction.
By combining token-level OT with motion-difference filtering, GRACE produces interpretable alignments between descriptive language and dynamic facial segments.
Extensive experiments on DFEW, FERV39k, and MAFW show that our method establishes a new SOTA.

\noindent\textbf{Our main contributions are summarized as follows:}
\begin{itemize}
 \item We design a granularity-aware text conditioning module CATE that selectively emphasizing emotion-relevant tokens for fine-grained language guidance.
 \item We propose a motion-difference based weighting mechanism that amplifies emotion-bearing dynamics while suppressing irrelevant motions.
 \item We formulate a token-level cross-modal alignment strategy using entropy-regularized Optimal Transport, enabling interpretable matching between semantic tokens and spatio-temporal patches, and achieving new SOTA results on three in-the-wild benchmarks.
\end{itemize}

\begin{figure*}
\includegraphics[width=7.125in]{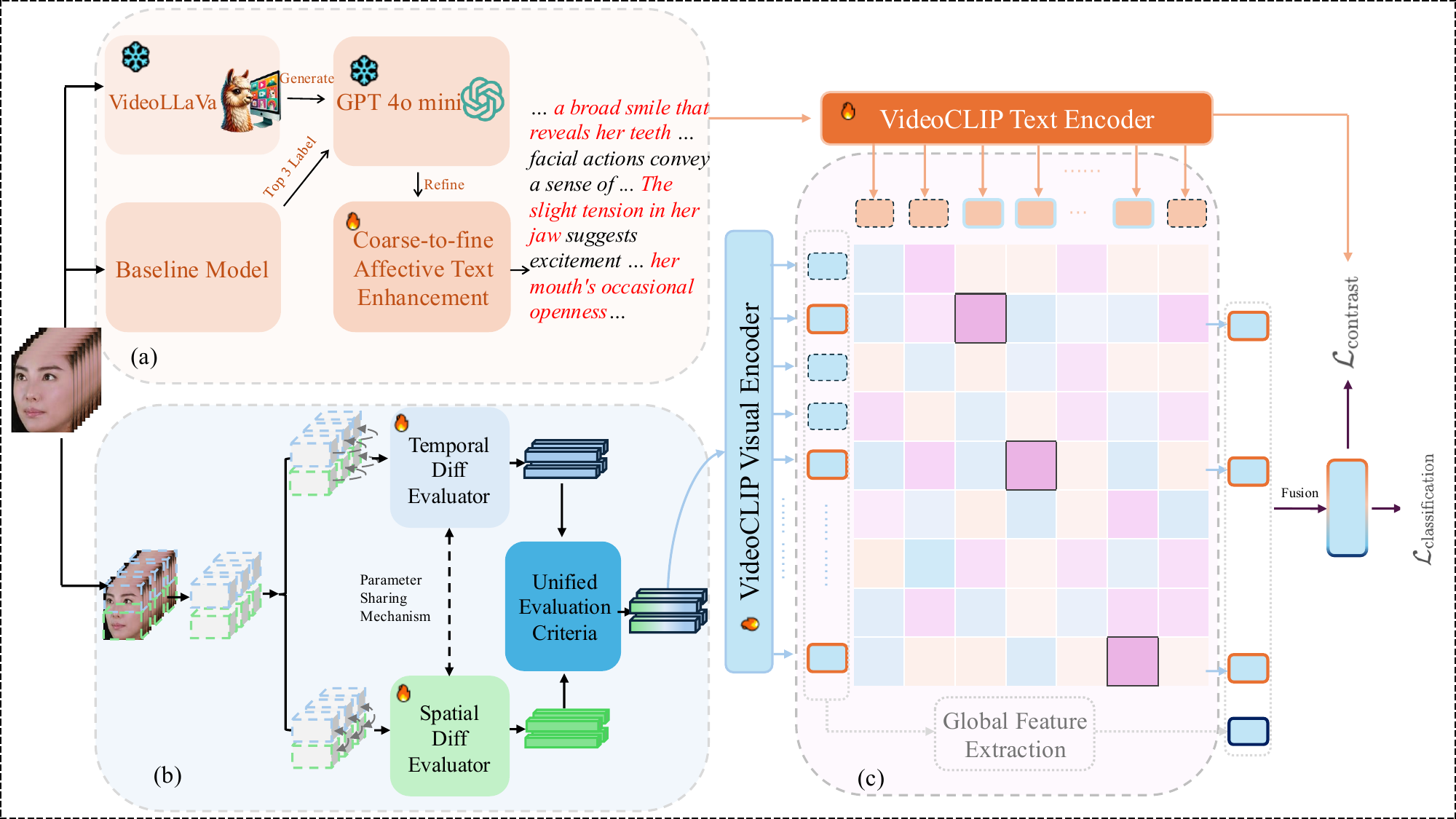}
\caption{Detailed architecture of the proposed GRACE framework. (a) \textbf{Coarse-to-fine Affective Text Enhancement
Module}: We first generate temporally-aligned, semantically rich descriptions from facial sequences using a VideoLLaVA-based pipeline enhanced by FACS-based prompt engineering and GPT-4o mini refinement. The resulting textual embeddings provide anatomically grounded semantic anchors for rare and ambiguous expressions. (b) \textbf{Motion-Aware Visual Representation Learning}: A Motion Difference Module calculates spatiotemporal dissimilarity matrices across facial regions, amplifying transient dynamics through unified evaluation criteria. This facilitates the identification of subtle motions critical for minority-class recognition. (c) \textbf{Optimal Transport-Based Ambiguity-Aware Matching}: The enriched visual and textual features are aligned using an optimal transport formulation, with dynamic cost matrices and entropy-controlled prompt buckets to balance ambiguity and long-tail biases. The final multimodal features are optimized via contrastive and classification losses.}
\label{fig:detailed_arch}
\end{figure*}

\section{Related Work}
\label{sec:related}

\subsection{Vision-Language Method for DFER}

The rise of large-scale vision-language pretraining~\cite{clip} has catalyzed semantic-aware DFER approaches by enabling the integration of linguistic priors into visual expression recognition. Initial adaptations like DFER-CLIP~\cite{DFER-CLIP} aligned video-level embeddings with class-level textual prompts (e.g., “\textit{a video of happiness}”), demonstrating improved generalization over purely visual baselines, especially under cross-domain and low-data regimes. Subsequent advances refined this paradigm along two major lines: (1) DK-CLIP~\cite{DK-CLIP} incorporated domain-specific affective knowledge via attribute-enhanced prompts, introducing auxiliary supervision grounded in emotion-related concepts (e.g., “furrowed brows” for anger); and (2) FineCLIPER~\cite{FineCLIPER} explored frame-level descriptions referencing Action Units (AUs), generating fine-grained textual labels to better guide frame-wise video encoding.

Meanwhile, methods like CEPrompt~\cite{13zhou2024ceprompt} from the static facial emotion recognition literature have also shown that incorporating concept-guided adapters into VLP frameworks enhances the expressiveness of visual representations, providing inspiration for DFER scenarios. Similarly, Yuan et al.~\cite{140} proposed a framework to describe facial expressions using language, bridging visual cues and emotional semantics through captioning pipelines, further highlighting the potential of multimodal learning for affective tasks.

However, these methods exhibit a critical architectural constraint: despite leveraging fine-grained textual descriptions (e.g., “raised eyebrows and parted lips”), they universally compress sentences into monolithic embedding vectors for video-level alignment. This design overlooks the \textit{compositional semantics} of language—failing to exploit word-level affective cues (e.g., “furrowed” vs. “wrinkled”) or phrase-level emotional salience. As a result, three key opportunities are lost: (1) associating transient facial actions with specific descriptive tokens, (2) leveraging linguistic hierarchies where certain terms (e.g., “snarl”) carry stronger discriminative power than others (e.g., “smile”), and (3) establishing spatially or temporally localized correspondences between visual and textual modalities~\cite{13zhou2024ceprompt,140}. This underutilization of linguistic granularity fundamentally limits the capacity of current frameworks to provide the fine-grained semantic guidance required for precise expression analysis.

\subsection{Granularity-Preserving Alignment in NLP and Vision-Language Tasks}

The benefit of retaining multi-level semantic structure has been extensively studied in NLP and multimodal research. A body of work demonstrates that retaining and selectively amplifying multi-granular cues—whether realised via explicit hierarchical encoders or via feature-level boosting mechanisms—can benefit downstream reasoning. In NLP (e.g., discourse parsing, coreference, relation extraction), keeping sentence-, phrase-, and token-level semantics improves the localisation of decisive clues \cite{liu2019hierarchical,hdt2024,bert}. Likewise, in vision–language modelling, methods that highlight phrase- or word-specific features guide attention to semantically relevant regions \cite{li2020hero,nguyen2019mtl,ye2022hmn}. These designs improve both interpretability and retrieval accuracy by leveraging fine-grained cross-modal correspondences.

Building on these insights, recent research has explored mechanisms that can support fine-grained alignment across modalities—not just through hierarchical encoding, but also via structured matching objectives. Among these, Optimal Transport~\cite{cuturi2013sinkhorn,montesuma2024recent} has emerged as a principled approach for aligning token-level representations between heterogeneous domains, offering a natural fit for modeling the cross-modal correspondence needed in granularity-aware settings.

Optimal transport originally proposed to model distances between probability distributions, has recently gained attention in diverse domains such as domain adaptation~\cite{Xu_Liu_Wang_Chen_Wang_2020}, clustering~\cite{Caron_Misra_Mairal_Goyal_Bojanowski_Joulin_2020}, document matching~\cite{Yu_Pang_Xu_Su_Dong_Wen}, and sequence alignment~\cite{Liu_Tekin_Coskun_Vineet_Fua_Pollefeys_2021}. Recent work has begun to apply token-level Optimal Transport to video–text alignment~\cite{chen2024and,lin2024multi}, demonstrating improved matching fidelity for fine-grained visual–linguistic correspondence. However, these methods are primarily developed for general-purpose video understanding tasks—such as action recognition or video retrieval—and have not been adapted to the unique demands of DFER, where emotion-relevant signals are both transient and spatially localized. 
Moreover, the token–patch matching strategies in prior work often assume explicit and salient motion cues, which may not hold in affective video settings where subtle expressions correspond to highly specific emotional descriptors. These subtle, semantically entangled patterns require alignment mechanisms that are not only temporally precise, but also linguistically grounded. 

\subsection{Challenges in Motion-Aware and Granular Alignment}

Beyond semantic limitations, effective DFER demands precise modeling of \textit{emotion-relevant motion} to differentiate subtle expression changes over time. Early optical flow-based methods~\cite{S2D,10zhao2016predicting} captured rigid motion patterns by analyzing frame-wise differences but were highly sensitive to non-expressive artifacts such as head rotations, eye blinks, or illumination variations. Landmark-based approaches~\cite{FineCLIPER,zhao2022spatial} introduced anatomical grounding through AU-informed facial tracking, offering more stable geometric features. However, these approaches often lack temporal modulation mechanisms to dynamically weigh AU intensity or suppress irrelevant motion segments.

To address these issues, frequency-domain techniques such as Freq-HD~\cite{Freq-HD} analyze multi-band temporal-spatial variations, allowing detection of periodic emotional signals. Yet, these methods often struggle with transient or subtle expressions that fall outside predictable frequency bands, limiting their granularity. Similarly, PPDN~\cite{10zhao2016predicting} and GCA-IAL~\cite{47li2023intensity} attempt to localize peak-expressive frames or modulate intensity-aware loss, but still admit interference from emotionally irrelevant movements, such as habitual blinking or lip twitches. Without explicit suppression mechanisms, such noise propagates into high-level recognition features, degrading model precision and interpretability.

This necessitates dual advancements: (1) motion modeling that selectively amplifies expressive facial movements while filtering parasitic activities, and (2) granular cross-modal alignment capable of linking visual dynamics with semantically meaningful descriptors.

Recent efforts in vision-language modeling have explored fine-grained video-text alignment techniques in domains such as action recognition and video retrieval~\cite{lin2024multi}, showing that token-level correspondence between modalities—implemented via cross-modal attention~\cite{tsai2019multimodal}, patch–word interaction~\cite{kim2021vilt, li2022blip}—has proved effective in action recognition and video retrieval. However, directly applying such alignment strategies to DFER remains nontrivial due to the unique characteristics of emotional expressions—such as subtle motion cues, inter-class ambiguities, and the presence of semantically neutral frames~\cite{FineCLIPER,47li2023intensity}.
Existing work on DFER either falls back to global pooling~\cite{FineCLIPER} or treats the caption as a single prompt~\cite{DK-CLIP, cliper}, leaving token-level alignment unexplored.

These challenges highlight the need for emotion-specific alignment mechanisms that not only ground visual features in linguistically rich supervision, but also adapt to the nuanced nature of dynamic facial expressions.

\section{Method}
In this section, we present our \textbf{\underline{G}ranular \underline{R}epresentation \underline{A}lignment for \underline{C}ross-modal \underline{E}motion recognition (\textsc{GRACE})} architecture, which addresses the challenge of DFER through a novel end-to-end multimodal architecture. Our framework innovatively combines motion dynamics analysis with semantic understanding to achieve more accurate and interpretable emotion recognition results.

\subsection{Overview}
\label{sec:overview}

We propose a novel multimodal framework named \textsc{GRACE}, which integrates semantic understanding and visual motion dynamics to address the challenges of accurate and interpretable DFER. The framework is designed to selectively identify emotionally salient spatiotemporal patterns by aligning fine-grained language semantics with motion-sensitive visual features.

As illustrated in Fig.~\ref{fig:detailed_arch}, \textsc{GRACE} consists of three key components: (1) a \textit{Coarse-to-fine Affective Text Enhancement module} that constructs detailed emotion descriptions incorporating emotion-descriptor phrase; (2) a \textit{motion-difference enhanced visual encoder} that captures subtle facial motion variations through dynamic weighting; and (3) a \textit{cross-modal alignment module} based on optimal transport, which establishes token-level correspondences between semantic units and video segments. Together, these components form a unified pipeline that filters irrelevant visual dynamics and enriches semantic interpretability.

Formally, given an input video sequence $\mathcal{V} = \{v_t\}_{t=1}^T \in \mathbb{R}^{T \times H \times W \times 3}$ and a corresponding descriptive text $\mathcal{D}$, the goal is to predict an emotion label $y \in \mathcal{Y}$, where $\mathcal{Y} = \{1, 2, \dots, C\}$ denotes the emotion category set. The overall process is formulated as:

\begin{equation}
    y = \mathcal{F}\left(\mathcal{A}\left(\mathcal{M}\left(\mathcal{E}_v(\mathcal{V})\right),\ \mathcal{P}\left(\mathcal{E}_t(\mathcal{D})\right)\right)\right)
\end{equation}

Here, $\mathcal{E}_v$ and $\mathcal{E}_t$ denote the visual and textual encoding modules, which extract spatiotemporal features and token-level semantic embeddings, respectively. $\mathcal{M}$ applies a motion-difference weighting mechanism to enhance dynamic expression cues within the visual modality, while $\mathcal{P}$ performs emotion-tuned semantic enhancement on the textual side to highlight emotion-relevant content. $\mathcal{A}$ computes a cross-modal alignment between tokens and visual segments via an entropy-regularized optimal transport solver. Finally, $\mathcal{F}$ denotes the classification head that integrates the aligned features to produce the final emotion prediction.

By tightly coupling semantic refinement and motion dynamics, our framework addresses two critical limitations of existing approaches: (i) the underutilization of fine-grained emotional semantics in text, and (ii) the lack of filtering mechanisms to suppress visually irrelevant dynamics. In the following sections, we detail each module in the proposed framework.

\begin{figure*}[!t]
\centering
\includegraphics[width=\linewidth, trim=0 42 0 60, clip]{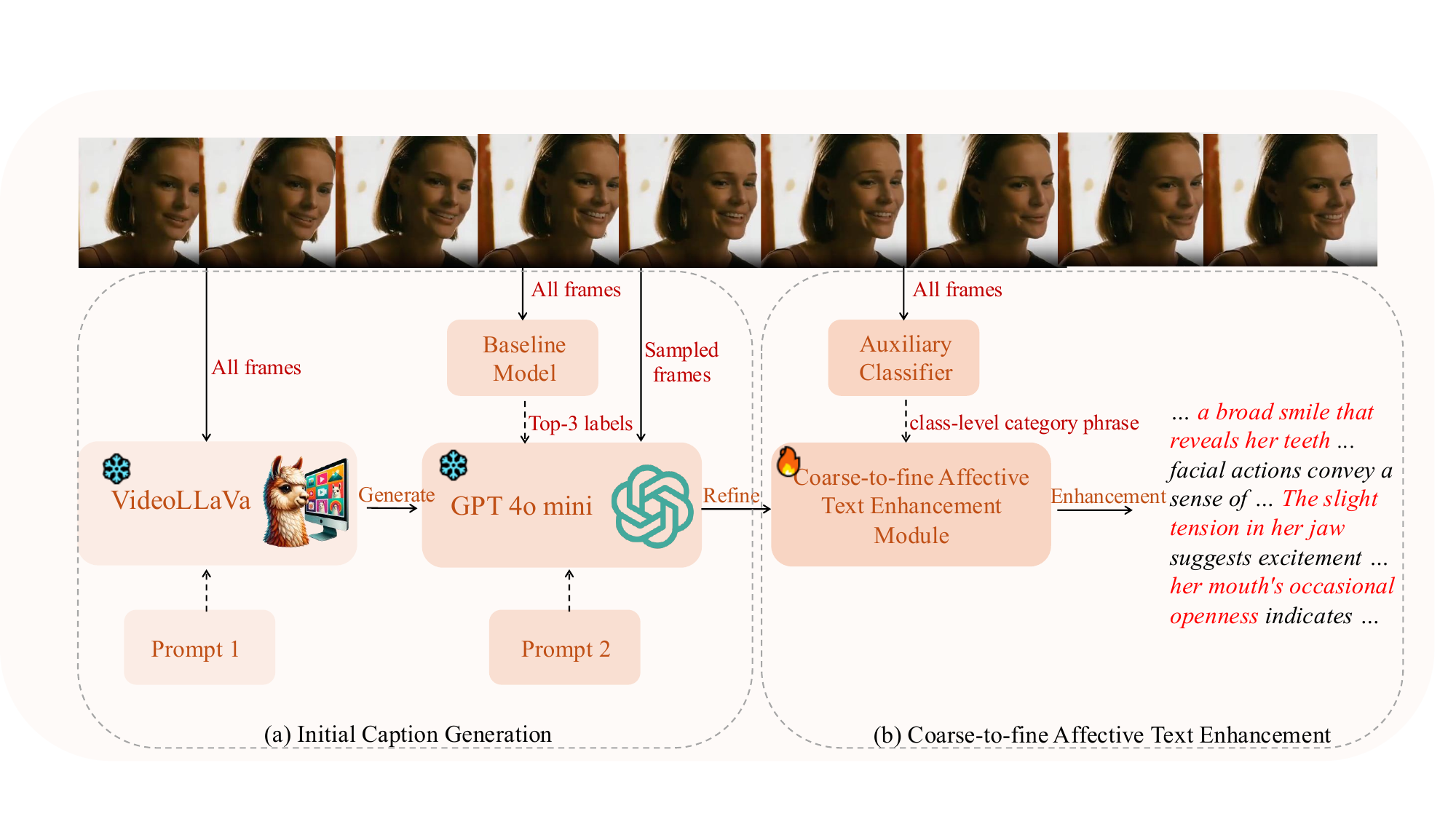}
\caption{Visualization of the CATE. This enhancement mechanism serves three key purposes: (1) it ensures semantic consistency with emotion-specific patterns, (2) it enables adaptive feature refinement at token level, and (3) it maintains the discriminative power of the original features through careful normalization. The enhanced token representations preserve both the fine-grained visual details from the original descriptions and the emotion-specific characteristics from the emotion-descriptor phrases.}
\label{fig:visualization_text}
\end{figure*}

\subsection{Category-Tuned Textual Enhancement}
\label{sec:textgen}

To incorporate fine-grained semantic guidance into the recognition process, we construct a textual description for each video that captures emotion-relevant facial behaviors. Unlike prior methods that treat emotion labels as discrete identifiers, we generate natural language descriptions that reflect the temporal and spatial dynamics of facial expressions. These descriptions serve not only as auxiliary supervision but also as semantic anchors for aligning with visual features at a token level.

\paragraph{Initial Caption Generation}
As illustrated in Fig.~\ref{fig:visualization_text}(a), we employ a pretrained vision-language model~\cite{Video-llava} to generate an initial natural language description $\mathcal{D}_{\text{init}}$ for each input video $\mathcal{V}$. The VLM encodes the video into a visual embedding and generates captions that describe observable facial movements, such as “the person slowly raises their eyebrows and opens their mouth.”
To further enhance the emotional granularity of these preliminary descriptions, we integrate a refinement process leveraging the GPT-4o mini model. Specifically, we first apply a baseline emotion recognition model to extract the top-3 predicted emotion categories from sampled frames, offering a coarse but semantically aligned emotional context. These labels serve as guiding signals for GPT-4o mini, enabling it to reconcile overlapping emotional cues and generate a refined caption that more accurately encapsulates nuanced affective states. This process helps reduce ambiguity in the emotional interpretation and ensures that the resulting description captures both fine-grained visual actions and their corresponding affective connotations.

\paragraph{Coarse-to-fine Affective Text Enhancement}
To improve the emotional relevance of the generated captions, we introduce a emotion-guided refinement module. As shown in Fig.~\ref{fig:visualization_text}(b), for each video, we first obtain its top-$k$ predicted emotion categories $\{c_1, c_2, \dots, c_k\}$ using a baseline visual classifier. These predicted categories are converted into emotion-descriptor phrase of the form “an emotion of \textit{[class]}”, These emotion-informed prompts are concatenated with the initial caption and passed through a lightweight LLM-based rewriter~\cite{raffel2020exploring}, resulting in a refined, emotion-aware description $\mathcal{D}_{\text{final}}$.

\begin{figure}[!t]
\centering
\includegraphics[width=\linewidth, trim=0 40 0 40, clip]{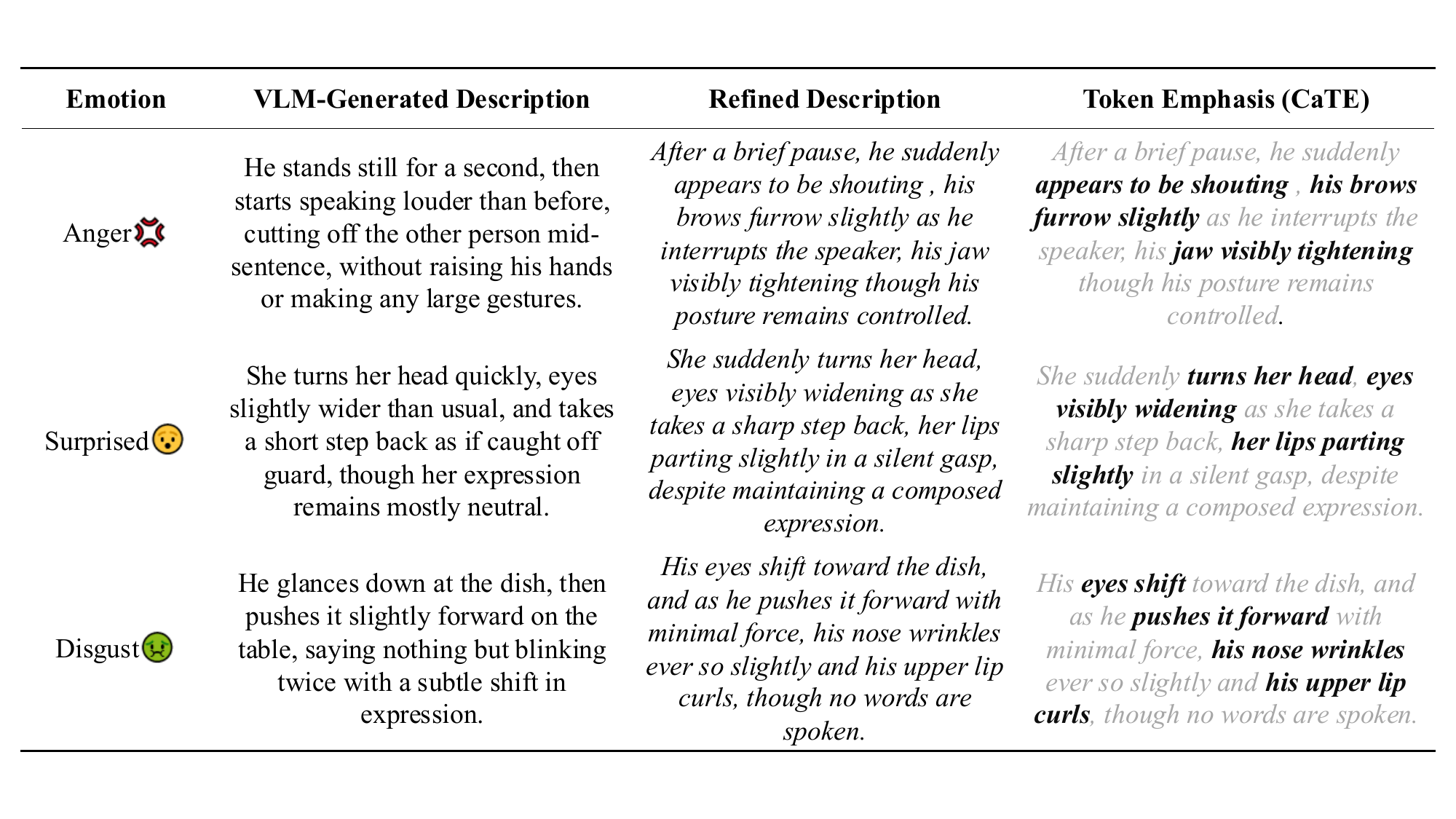}
\caption{Comparison of textual representations across three stages of our textual enhancement pipeline. 
}
\label{fig:text_granularity}
\end{figure}

This enhancement process introduces two critical benefits: (1) it injects explicit emotional semantics into the textual input, strengthening its alignment with category-level meaning; and (2) it encourages the model to emphasize action units or facial behaviors that are discriminative across similar emotion categories, thereby reducing ambiguity. Importantly, the final refined description is retained as a sequence of tokens, preserving its internal semantic structure for token-level alignment in subsequent modules. Fig.~\ref{fig:text_granularity} illustrates how this multi-stage textual enhancement pipeline improves emotional clarity. The first column shows initial captions generated by VLM from video input. These captions are then refined to produce more detailed emotion-relevant descriptions (middle). While the base captions may appear ambiguous, our CATE module highlights semantically relevant tokens—such as facial action descriptions—that play a critical role in distinguishing affective intent. By operating within the natural language space, this method preserves contextual completeness while enabling targeted emphasis aligned with emotion categories.

\paragraph{Token Representation}
The refined caption $\mathcal{D}_{\text{final}}$ is encoded into a sequence of token embeddings $\mathcal{T} = \{t_1, t_2, \dots, t_L\}$ using a pretrained textual encoder~\cite{xu2021videoclip}. These embeddings are passed into the alignment module alongside visual representations, allowing the model to establish fine-grained cross-modal correspondences.

This module addresses the first major limitation discussed in Section~\ref{sec:intro}: although fine-grained text can be generated, existing methods rarely exploit its internal semantic cues. By injecting class-level priors and preserving token-level structure, our approach enables a more expressive and discriminative use of textual guidance.

\subsection{Motion-Aware Visual Representation Learning}
\label{sec:visual}

To effectively capture emotion-relevant facial dynamics, we introduce a motion-difference based visual enhancement module that identifies salient spatiotemporal regions within videos. Unlike approaches that rely solely on frame-level semantics or fuse handcrafted motion cues (e.g., optical flow or facial landmarks), our method emphasizes learning motion sensitivity directly from visual features. This design avoids incorporating modality-specific biases and reduces preprocessing overhead, while enabling fine-grained focus on emotionally discriminative movements.

\paragraph{Feature Extraction}
We adopt VideoMAE~\cite{tong2022videomae} as the visual encoder $\mathcal{E}_v(\cdot)$ to obtain dense spatiotemporal representations and follow the MAE-DFER setting~\cite{MAE-DFER}. Given an input video $\mathcal{V} = \{v_t\}_{t=1}^T$, the encoder outputs a feature tensor $\mathcal{X} \in \mathbb{R}^{T' \times H' \times W' \times D}$, where $T'$ denotes the number of temporal segments, $H' \times W'$ represents spatial patches, and $D$ is the feature dimension. This representation preserves both temporal dynamics and spatial layout, providing a rich basis for identifying subtle facial motion.

\paragraph{Temporal and Spatial Motion Scoring}
To highlight regions and segments associated with facial expression changes, we compute motion saliency scores by applying inter-frame differencing to the extracted features. For each spatial location $(h, w)$ and time index $t$, we compute:

\begin{equation}
    \Delta\mathcal{X}_{t}^{h,w} = \left\| \mathcal{X}_{t}^{h,w} - \mathcal{X}_{t-1}^{h,w} \right\|_2
\end{equation}

This difference quantifies the magnitude of change in local visual features across adjacent time steps. Aggregating over channels yields a saliency heatmap that reflects motion intensity at each patch. Intuitively, regions with strong inter-frame variation are more likely to correspond to expression-related facial movements.

\paragraph{Dynamic Feature Weighting}
We normalize the computed motion scores to obtain a spatiotemporal attention map $\mathcal{W} \in \mathbb{R}^{T' \times H' \times W'}$, which is then used to reweight the original feature tensor:

\begin{equation}
    \widetilde{\mathcal{X}}_{t}^{h,w} = \mathcal{W}_{t}^{h,w} \cdot \mathcal{X}_{t}^{h,w}
\end{equation}

The resulting motion-weighted features $\widetilde{\mathcal{X}}$ selectively amplify facial regions undergoing significant expressive changes, while suppressing irrelevant dynamics (e.g., eye blinks, head tilts, or background motion). This enhances the model’s ability to focus on emotion-specific variations without introducing handcrafted modalities.

These refined features are subsequently passed to the cross-modal alignment module (Section~\ref{sec:alignment}), enabling token-level semantic matching with the textual descriptions. In this way, our motion-difference mechanism provides the visual pathway with fine-grained sensitivity to dynamic emotional signals, directly addressing the second limitation outlined in Section~\ref{sec:intro}.

\subsection{Token-Level Cross-Modal Alignment via Optimal Transport}
\label{sec:alignment}

To bridge the semantic gap between the refined textual descriptions and the motion-enhanced visual features, we adopt a token-level cross-modal alignment strategy based on optimal transport. This module enables precise matching between emotion-relevant words and spatiotemporal facial regions, allowing the model to focus on localized dynamic cues that correspond to specific semantic units.

Inspired by recent success in fine-grained video-text matching~\cite{lin2024multi}, we build upon differentiable Optimal Transport formulations~\cite{cuturi2013sinkhorn} to enable token-level soft alignment between temporal and textual embeddings.

\textit{However, DFER presents unique challenges}: high inter-class ambiguity (e.g., anger vs. disgust), low motion contrast in subtle expressions, and frequent occurrence of neutral frames~\cite{FineCLIPER,47li2023intensity} make direct OT adaptation prone to failure. Naive OT implementations often match ambiguous frames with emotionally charged text tokens, simply due to spatial similarity or lighting artifacts, introducing semantic misalignment.

Unlike prior OT-based video–text aligners that assume salient, large-scale motions, we adapt the transport cost with emotion-aware token weights and motion-difference filtering, explicitly tackling the semantic ambiguity and subtle local movements that distinguish DFER from generic action recognition.

Thus, successful OT-based fusion in DFER requires \textit{domain-specific refinements}—including emotion-aware text generation that emphasizes discriminative lexical choices (e.g., “tensed jaw” vs. “mouth closed”) and motion-saliency filtering modules that enhance emotionally salient segments while down-weighting generic dynamics. Our framework addresses this gap through a CATE and a motion-difference-based visual enhancement module designed specifically for dynamic facial expressions.

\paragraph{Input Representation}
Let $\mathcal{T} = \{t_1, t_2, \dots, t_L\} \in \mathbb{R}^{L \times d}$ denote the sequence of $L$ textual token embeddings obtained from the refined description (Section~\ref{sec:textgen}), and let $\widetilde{\mathcal{X}} = \{x_1, x_2, \dots, x_N\} \in \mathbb{R}^{N \times d}$ represent the set of $N$ motion-weighted visual patch embeddings from the enhanced feature map (Section~\ref{sec:visual}). Both are projected into a shared latent space using pretrained VideoCLIP encoders.

\paragraph{Cost Matrix and Transport Plan}
We compute a pairwise cost matrix $\mathbf{C} \in \mathbb{R}^{L \times N}$, where each element $C_{i,j}$ measures the dissimilarity between textual token $t_i$ and visual token $x_j$, typically defined as:

\begin{equation}
    C_{i,j} = 1 - \frac{t_i^\top x_j}{\|t_i\| \cdot \|x_j\|}
\end{equation}

This cosine distance captures semantic dissimilarity. We then solve for the optimal transport plan $\mathbf{T}^* \in \mathbb{R}^{L \times N}$ by minimizing the total transport cost with entropy regularization:

\begin{equation}
    \mathbf{T}^* = \arg\min_{\mathbf{T} \in \Pi(\mathbf{a}, \mathbf{b})} \langle \mathbf{T}, \mathbf{C} \rangle - \lambda H(\mathbf{T})
\end{equation}

Here, $\Pi(\mathbf{a}, \mathbf{b})$ denotes the set of doubly stochastic matrices with prescribed marginals $\mathbf{a}$ and $\mathbf{b}$ (typically uniform distributions over text and visual tokens), and $H(\mathbf{T})$ is the entropy of the transport plan. This optimization is solved efficiently using the Sinkhorn algorithm~\cite{cuturi2013sinkhorn}.

\paragraph{Cross-Modal Fusion and Objective}
The resulting transport matrix $\mathbf{T}^*$ serves as a soft alignment map, indicating how each word attends to each visual region. We use it in two ways: (1) to compute a transport-weighted fused representation for classification, and (2) to supervise the alignment using a cross-modal contrastive loss. Specifically, the aligned feature pairs $(t_i, x_j)$ weighted by $T_{i,j}^*$ are encouraged to be closer for matching pairs and repelled otherwise.

This module enables our model to not only localize emotion-relevant visual features but also interpret them in light of specific textual semantics. By aligning text and video at the token level, we ensure that subtle facial expressions—such as eyebrow raises or mouth tension—are directly grounded in emotional language cues, completing the semantic-visual loop introduced in earlier stages.

\subsection{Loss Functions}

To effectively train our multimodal emotion recognition framework, we design a comprehensive multi-task learning strategy that combines weighted focal loss, supervised contrastive learning, and auxiliary classification objectives. Each loss component is carefully designed to address specific challenges in emotion recognition.

\subsubsection{Multi-Task Learning Objectives} 

Our training objective consists of three complementary loss terms:

The primary objective employs a weighted focal loss to address class imbalance and hard example mining in emotion recognition:

\begin{equation}
    \mathcal{L}_{focal} = -\sum_{i=1}^{N}\sum_{c=1}^{C} w_c\alpha_t(1-p_t)^\gamma y_c^{(i)}\log(p_c^{(i)})
\end{equation}

where $w_c$ is the class-specific weight, $p_t$ is the model's estimated probability for the target class, $\gamma$ is the focusing parameter, and $\alpha_t$ is the balancing factor. This loss is specifically designed to handle the inherent class imbalance in emotion datasets by assigning higher weights to minority classes and focusing more on hard examples through the modulating factor $(1-p_t)^\gamma$.

To enhance the alignment between visual and textual representations while preserving emotion-specific characteristics, we employ a supervised contrastive loss with mixup augmentation:

\begin{equation}
    \mathcal{L}_{supcon} = \sum_{i=1}^{N}\frac{-1}{|P(i)|}\sum_{p\in P(i)}\log\frac{\exp(f_v^{(i)}\cdot f_t^{(p)}/\tau)}{\sum_{j=1}^{2N}[\![ j{\neq}i ]\!]\exp(f_v^{(i)}\cdot f_t^{(j)}/\tau)}
\end{equation}

where $P(i)$ is the set of positive pairs for the $i$-th sample after mixup augmentation, $\tau$ is the temperature parameter, and $f_v$, $f_t$ are the normalized visual and textual features respectively. This contrastive learning objective helps to learn discriminative features by pulling together representations of the same emotion while pushing apart different emotions in the embedding space, thereby enhancing the model's ability to distinguish subtle emotional differences.

To guide the text feature enhancement module with emotion-informed prompts, we introduce an auxiliary classification loss:

\begin{equation}
    \mathcal{L}_{aux} = -\sum_{i=1}^{N}\sum_{c=1}^{C} y_c^{(i)}\log(p_c^{(i)})
\end{equation}

This auxiliary loss serves as a regularizer for the text feature enhancement module, ensuring that the enhanced text features maintain their emotion-discriminative properties while being aligned with the visual modality. It helps to preserve the semantic information of emotion categories during the feature enhancement process.

The total loss is a weighted combination of these components:

\begin{equation}
    \mathcal{L}_{total} = \lambda_1\mathcal{L}_{focal} + \lambda_2\mathcal{L}_{supcon} + \lambda_3\mathcal{L}_{aux}
\end{equation}

where $\lambda_1$, $\lambda_2$, and $\lambda_3$ are carefully tuned hyperparameters that balance the contribution of each loss term. This weighted combination ensures that each learning objective contributes appropriately to the overall optimization process, leading to better model convergence and performance.

\section{Experiments}

\subsection{Experimental Settings and Implementation Details}

\subsubsection{Datasets}

We perform experiments on two widely used benchmarks for dynamic facial expression recognition: \textbf{DFEW}, \textbf{FERV39k} and \textbf{MAFW}. Details of each dataset are as follows:

\textbf{DFEW} is a popular and widely adopted dynamic facial emotion recognition dataset containing 16,372 video samples labeled with 7 basic emotional categories (happiness, sadness, neutral, anger, surprise, disgust, and fear). Videos are extracted from over 1,500 movies and annotated by 10 professionally trained annotators, making it representative of real-world, in-the-wild conditions. Critically, DFEW exhibits a pronounced long-tail distribution, where the majority class, happiness, has 2,824 samples compared to only 87 samples in the minority class, disgust. This severe imbalance provides a natural testbed to rigorously assess our model’s robustness to dataset biases and its ability to handle semantically sparse classes.

\textbf{FERV39k} is currently the largest publicly available dynamic facial expression recognition dataset, comprising 38,935 video clips across 22 representative scenes grouped into 4 distinct scenarios. Each video clip has been rigorously annotated by 30 professional annotators, covering the same 7 basic emotion categories as DFEW. Notably, FERV39k has a relatively balanced emotion distribution compared to DFEW, making it ideal for evaluating cross-dataset generalization. The dataset is officially split into a training set (80\%) and a testing set (20\%). We leverage FERV39k’s balanced characteristics to validate our method’s capability to generalize beyond the biases inherent in training datasets.

\textbf{MAFW} is a multimodal affective dataset collected under unconstrained conditions. It consists of 10,045 video clips annotated with 11 compound emotions, including contempt, anxiety, helplessness, disappointment, and seven basic emotions. In this work, only the video modality is considered, and experiments are conducted on 9,172 single-labeled video clips. Following the protocol in the original study~\cite{liu2022mafw}, five-fold cross-validation is employed for evaluation.

\subsubsection{Preprocessing}

To ensure high-quality facial input for motion-aware modeling, we uniformly sample 16 frames per video with a fixed stride and resize each to $224 \times 224$ via center cropping. Data augmentation includes scale jittering, horizontal flipping, and erasing-based regularization to improve generalization. Class imbalance is addressed via inverse-root weighting based on sample frequency. At evaluation, we apply multi-segment, multi-crop testing to enhance prediction robustness.

\subsubsection{Implementation Details}

\textbf{Pretrained models.}
To ensure fair comparison with existing methods, we utilize publicly available pretrained weights for all base models. Specifically, our framework integrates three pretrained components:
(1) A motion-aware visual backbone based on VideoMAE pretrained on VoxCeleb2, following the MAE-DFER~\cite{MAE-DFER} setup. This configuration facilitates robust facial feature extraction and efficient GPU resource utilization via LGIFormer.
(2) A cross-modal encoder based on VideoCLIP, pretrained on the large-scale HowTo100M dataset~\cite{xu2021videoclip}, enabling effective video-text alignment.
(3) Additionally, coarse textual descriptions are generated by VideoLLaVA, pretrained on LLaVA-Instruct-150K~\cite{Video-llava}, supplemented with filtered image-text and video-text pairs from LAION-CC-SBU~\cite{li2022blip}, CC3M~\cite{sharma2018conceptual}, and WebVid~\cite{bain2021frozen}. These initial descriptions are then further refined by the OpenAI GPT-4o mini model to produce fine-grained textual descriptions.
These pretrained components provide strong priors for visual representation learning, multimodal fusion, and text generation.

\textbf{Training details.}
We utilize a batch size of 64 and train for 30 epochs to ensure model convergence, while maintaining reproducibility through a fixed random seed of 3407 across all experimental runs. For regularization, we implement a multi-level strategy with dropout rate of 0.0, attention dropout rate of 0.0, and drop path rate of 0.1 to prevent overfitting, while employing GELU activation functions which demonstrate superior performance in Vision Transformer architectures. The optimization is performed using AdamW optimizer with learning rate of 1e-3, weight decay of 0.05, momentum of 0.9, and epsilon of 1e-8, complemented by gradient clipping to prevent gradient explosion. We adopt a cosine annealing scheduler with initial learning rate of 1e-3, minimum learning rate of 1e-6, and warmup period of 5 epochs, along with layer-wise learning rate decay (decay rate of 0.75) to accommodate the parameter distribution of pre-trained models. The experiments are conducted using PyTorch 1.10 with CUDA 11.3.

\subsubsection{Evaluation Metrics}

Consistent with previous works, we evaluate our model using two standard metrics: Weighted Average Recall (WAR) and Unweighted Average Recall (UAR).

Notably, UAR is especially critical for DFER tasks in clinical scenarios, as it treats all emotion classes equally regardless of sample distribution, thus reflecting genuine performance in practical settings where minority classes (e.g., disgust, fear) are often diagnostically significant yet easily overlooked by biased metrics such as WAR~\cite{metze2010emotion, gosztolya2018posterior}.

\subsection{Comparison with SOTA Methods}

We compare the performance of our proposed GRACE framework against recent SOTA methods on three widely used datasets: DFEW, FERV39k, and MAFW. The comparison results are summarized in 
Table~\ref{tab:comparison_sota}.

On DFEW, GRACE achieves the highest UAR of 68.94\% and a WAR of 76.25\%, surpassing both visual-only methods such as S2D~\cite{S2D}, and M3DFEL~\cite{M3DFEL}, as well as visual-language-based methods like FineCLIPER~\cite{FineCLIPER} and A$^3$lign-DFER~\cite{a3ling}. Similarly, on FERV39k, GRACE attains a UAR of 49.12\% and a WAR of 54.63\%, outperforming strong baselines including MAE-DFER~\cite{MAE-DFER} and DFER-CLIP~\cite{DFER-CLIP}. On MAFW, GRACE also demonstrates competitive performance with a UAR of 45.09\% and a WAR of 58.25\%, establishing a new benchmark across multiple datasets.

Beyond numerical improvements, GRACE’s design contributes to performance gains in specific, structurally explainable ways. The CATE enables the model to focus on semantically critical emotion descriptors, leading to more accurate recognition of nuanced expressions like fear and disgust, which are often underrepresented and semantically ambiguous. Meanwhile, our motion-aware visual filtering mechanism effectively suppresses non-expressive dynamics, allowing the model to isolate emotionally salient cues in a sequence. Finally, the use of token-level optimal transport for cross-modal alignment ensures that emotionally relevant words are precisely matched with expressive facial regions, contributing to both improved accuracy and model interpretability. This synergy is particularly evident on long-tail classes and ambiguous expressions, where previous methods typically fail to generalize.

\begin{table*}[!t]
\label{tab:three_datasets}
\caption{Comparisons of our GRACE with the SOTA DFER methods on DFEW, FERV39k, and MAFW. Baseline results are directly extracted from~\cite{MAE-DFER}. The best results are highlighted in \textbf{bold}, and the second-best \underline{underlined}. 
${*}$: Baseline with OT Alignment; ${\dagger}$: Baseline with CATE; ${\ddagger}$: Baseline with MotionDiff.}
\label{tab:comparison_sota}
\centering
\renewcommand{\arraystretch}{1.1}
\begin{tabular}{lcccccccc}
\toprule
\multirow{2}{*}{\textbf{Method}} 
 & & \multicolumn{2}{c}{\textbf{DFEW}} & \multicolumn{2}{c}{\textbf{FERV39k}} & \multicolumn{2}{c}{\textbf{MAFW}} \\
\cmidrule(lr){3-4} \cmidrule(lr){5-6} \cmidrule(lr){7-8}
& & UAR (\%) & WAR (\%) & UAR (\%) & WAR (\%) & UAR (\%) & WAR (\%) \\
\midrule
\multicolumn{8}{l}{\textbf{\textit{Single Visual Modality}}} \\
EC-STFL~\cite{jiang2020dfew} & & 45.35 & 56.51 & - & - & - & - \\
Former-DFER~\cite{former} & & 53.69 & 65.70 & 37.20 & 46.85 & 31.16 & 43.27 \\
CEFLNet~\cite{liu2022clip} & & 51.14 & 65.35 & - & - & - & - \\
NR-DFERNet~\cite{nr-dfernet} & & 54.21 & 68.19 & 33.99 & 45.97 & - & - \\
Freq-HD~\cite{Freq-HD} & & 46.85 & 55.68 & 33.07 & 45.26 & - & - \\
LOGO-Former~\cite{logo} & & 54.21 & 66.98 & 38.22 & 48.13 & - & - \\
IAL~\cite{ial} & & 55.71 & 69.24 & 35.82 & 48.54 & - & - \\
T-MEP~\cite{t-mep} & & 57.16 & 68.85 & - & - & 39.37 & 52.85 \\
M3DFEL~\cite{M3DFEL} & & 56.10 & 69.25 & 35.94 & 47.67 & - & - \\
SVFAP~\cite{svfap}  & & 62.83 & 74.27 & 42.14 & 52.29 & 41.19 & 54.28 \\
MAE-DFER~\cite{MAE-DFER}  & & 63.41 & 74.43 & 43.12 & 52.07 & 41.62 & 54.31 \\
S2D~\cite{S2D} & & 65.45 & 74.81 & 43.97 & 46.21 & 43.40 & 52.55 \\
\midrule
\multicolumn{8}{l}{\textbf{\textit{Visual-Language Modality}}} \\
CLIPER~\cite{cliper} &  & 57.56 & 70.84 & 41.23 & 51.34 & - & - \\
DFER-CLIP~\cite{DFER-CLIP} & & 59.61 & 71.25 & 41.27 & 51.65 & 39.89 & 52.55 \\
EmoCLIP~\cite{emoclip} & & 58.04 & 62.12 & 31.41 & 36.18 & 34.24 & 41.46 \\
A$^3$lign-DFER~\cite{a3ling} & & 64.09 & 74.20 & 41.87 & 51.77 & 42.07 & 53.24 \\
HiCMAE~\cite{hicmae} & & 63.76 & 75.01 & - & - & 42.65 & 56.17 \\
FineCLIPER~\cite{FineCLIPER} & & 65.98 & \underline{76.21} & 45.22 & \underline{53.98} & \underline{45.01} & \underline{56.91} \\
\rowcolor{blue!5}
GRACE$^{*}$ & & 61.61 & 75.05 & 43.97 & 52.81 & 43.49 & 54.25 \\
\rowcolor{blue!5}
GRACE$^{*}$$^{\dagger}$ & & 63.58 & 74.97 & 44.95 & 52.34 & 43.97 & 55.12 \\
\rowcolor{blue!5}
GRACE$^{*}$$^{\ddagger}$ & & \underline{67.44} & 74.20 & \underline{47.96} & 53.45 & 44.78 & 56.16 \\
\rowcolor{blue!10}
GRACE$^{*}$$^{\dagger}$$^{\ddagger}$ & & \textbf{68.94} & \textbf{76.25} & \textbf{49.12} & \textbf{54.63} & \textbf{45.09} & \textbf{58.25} \\
\bottomrule
\end{tabular}
\end{table*}

As indicated in Table~\ref{tab:comparison_sota}, the full GRACE model consistently outperforms its ablated versions, confirming the importance and complementary roles of each proposed module. These comprehensive comparisons demonstrate the superiority and robustness of GRACE across different datasets, evaluation metrics, and emotion categories.

\section{Ablation Study}

We conduct comprehensive ablation studies to evaluate the individual contribution of each proposed module: OT alignment, textual enhancement with FACS refinement, and visual enhancement through the MotionDiff module. 

For the overall ablation (Table~\ref{tab:overall_ablation}), we progressively integrate each component into the baseline model and observe consistent improvements in both UAR and WAR metrics. To further validate the effectiveness of each module independently, we perform fine-grained ablations (Tables~\ref{tab:ot_ablation}-\ref{tab:visual_ablation}), where we selectively remove or replace a single module while keeping other components activated. 

The results demonstrate that each module contributes positively to the final performance, with the full model achieving the best overall scores. These results confirm that both the textual and visual enhancements, along with the OT-based alignment, are crucial to the system's superior performance.

\textbf{Progressive Integration under the OT Backbone.} 
We note that the modules introduced in GRACE—namely CATE and the motion-difference visual enhancer—are designed to work in conjunction with the OT alignment backbone. Specifically, the CATE module enhances textual prompts to produce semantically discriminative token-level anchors, which are meaningful only when paired with a transport-based cost formulation. Without OT, these refined tokens would lack a mechanism for effective visual grounding, rendering the CATE output functionally disconnected from the visual stream.

Similarly, although our motion-difference visual enhancement module can theoretically operate without OT, its core purpose is to improve the expressiveness of the patch-level dynamics that OT subsequently aligns with emotion-bearing words. Since these modules are not standalone classifiers but instead act as \textbf{modality-specific refinements to guide cross-modal alignment}, we treat OT alignment as a foundational component of the GRACE framework.

\begin{table}[ht]
\centering
\caption{Overall ablation study showing the performance impact of progressively integrating each module into the baseline model. The results were obtained on Fold 1 of the DFEW dataset.}
\begin{tabular}{lcc}
\toprule
\textbf{Model} & \textbf{UAR (\%)} & \textbf{WAR (\%)} \\
\midrule
Baseline & 61.04 & 73.81 \\
Baseline + OT Alignment & 61.61 & 75.05 \\
Baseline + OT + CATE & 63.58 & 74.97 \\
Baseline + OT + Visual Enhancement & 67.44 & 74.20 \\
Full Model (GRACE) & \textbf{68.42} & \textbf{76.29} \\
\bottomrule
\end{tabular}
\label{tab:overall_ablation}
\end{table}

As such, Table~\ref{tab:overall_ablation} focuses on a realistic, progressive integration path: starting from the baseline with OT alignment, and incrementally adding textual and visual enhancements that leverage the alignment backbone. This not only reflects the actual working conditions of GRACE, but also avoids misleading interpretations that would arise from testing modules in unnatural configurations.

Each component contributes positively, with the full GRACE model delivering the highest UAR (68.42\%) and WAR (76.29\%) scores, validating the complementary benefits of the proposed modules.

\begin{table}[ht]
\centering
\caption{Joint ablation study on text granularity and alignment strategy (UAR/WAR). All experiments were conducted on Fold 1 of the DFEW dataset.}
\begin{tabular}{lcc}
\toprule
\textbf{Alignment$\downarrow$~/~Text$\rightarrow$} & Sentence-level token & Fine-grained token \\
\midrule
Cosine Similarity Matching & 60.92/73.39 & 61.10/73.47 \\ 
Cross-Attention Matching & 60.08/73.26 & 60.65/74.33 \\ 
OT Alignment (Full Model) & 66.21/74.54 & \textbf{68.42/76.29} \\
\bottomrule
\end{tabular}
\label{tab:ot_ablation}
\end{table}
\textbf{Joint Ablation of Text Granularity and Alignment Strategy.} To disentangle the individual and joint effects of textual granularity and alignment strategy, we conduct a ablation shown in Table~\ref{tab:ot_ablation}. Along the horizontal axis, we vary the level of tokenization in the generated captions—comparing sentence-level tokens with fine-grained emotion tokens. Vertically, we compare three matching strategies: cosine similarity, cross-attention, and our entropy-regularized optimal transport.

We observe that both components contribute to performance independently. For example, switching from sentence-level to fine-grained tokens under OT yields a +2.21\% UAR gain, highlighting the benefit of preserving fine-grained emotional semantics. Similarly, under fine-grained inputs, using OT instead of cosine matching brings a +7.32\% UAR gain, validating the effectiveness of our alignment formulation.

Crucially, the full model (fine-grained token + OT alignment) outperforms all other combinations by a significant margin (68.42\% UAR vs. 60.92\% for the weakest setting), suggesting a synergistic interaction between precise linguistic anchoring and regularized alignment. These results confirm that GRACE benefits not from a single component, but from a carefully coordinated design across modalities.

\begin{table}[!ht]
\centering
\caption{Ablation study on the textual enhancement module. Other components remain activated. The results were obtained on Fold 1 of the DFEW dataset.}
\begin{tabular}{lcc}
\toprule
\textbf{Setting} & \textbf{UAR (\%)} & \textbf{WAR (\%)} \\
\midrule
w/o emotion-descriptor phrase & 67.84 & 75.38 \\ 
w/o Top-3 labels guidance & 66.27 & 74.28 \\
Full Model & \textbf{68.42} & \textbf{76.29} \\
\bottomrule
\end{tabular}
\label{tab:textual_ablation_module}
\end{table}

\textbf{Textual Enhancement Ablation.}
To assess the contribution of our semantic guidance strategy, we conduct ablations on two core components of the textual generation module: emotion-descriptor phrase and Top-3 label guidance. As shown in Table~\ref{tab:textual_ablation_module}, removing the emotion-descriptor phrase results in consistent performance degradation, confirming the importance of integrating emotion-specific semantic phrases into the text descriptions. This component enables the model to emphasize discriminative action units tied to each emotion class, thus providing fine-grained linguistic cues that improve recognition.

Similarly, discarding the Top-3 label guidance leads to a further decline in both UAR and WAR. Without this mechanism, the generated descriptions lack directional supervision and may include less relevant content. This highlights the role of classification-aware guidance in narrowing the semantic focus of the generated text, especially for compound or ambiguous emotions. Together, these two enhancements form a synergistic mechanism for semantically grounded expression understanding.

\begin{table}[ht]
\centering
\caption{Ablation study on the visual enhancement module. Other components remain activated. The results were obtained on Fold 1 of the DFEW dataset.}
\begin{tabular}{lcc}
\toprule
\textbf{Setting} & \textbf{UAR (\%)} & \textbf{WAR (\%)} \\
\midrule
w/o Motion Difference Module & 63.58 & 74.97 \\
Spatial Difference Only & 66.17 & 72.66 \\
Temporal Difference Only & 66.98 & 73.94 \\
Spatial + Temporal (Full Model) & \textbf{68.42} & \textbf{76.29} \\
\bottomrule
\end{tabular}
\label{tab:visual_ablation}
\end{table}
\textbf{Visual Enhancement Ablation.}
To evaluate the role of motion-sensitive modeling, we ablate the visual enhancement mechanism that dynamically highlights expressive regions based on inter-frame differences. As shown in Table~\ref{tab:visual_ablation}, removing this component significantly reduces performance, particularly in UAR, indicating the model's diminished ability to isolate emotionally salient temporal patterns.

Further analysis reveals that using only spatial or only temporal difference signals yields limited gains, whereas their combination leads to the best performance. This supports our hypothesis in the Introduction: that effective dynamic facial expression recognition hinges on suppressing irrelevant facial dynamics—such as head tilts or blinking—while enhancing subtle, emotion-relevant motion cues. By incorporating both spatial and temporal motion difference maps, our framework selectively attends to regions exhibiting discriminative expression patterns, thereby improving both accuracy and robustness.

\begin{figure}[!t]
\label{fig:tsne_visualization}
\centering
\includegraphics[width=\linewidth, trim=15 52 15 0, clip]{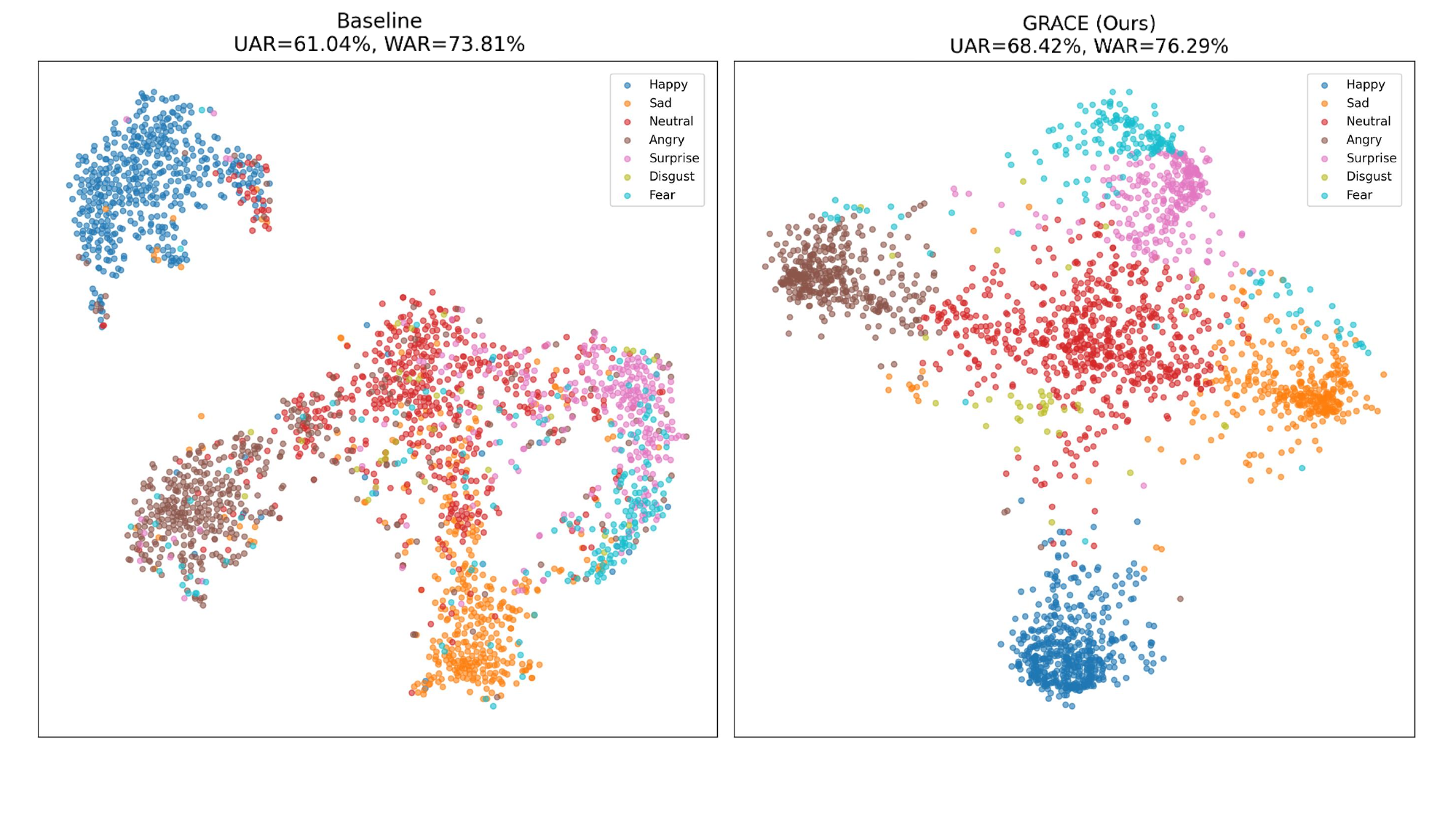}
\caption{t-SNE visualization of learned emotion representations for the Baseline method (left) and our proposed GRACE (right). Each point represents a sample, colored by its ground-truth emotion class. Compared to the baseline, GRACE exhibits more compact intra-class clusters and clearer inter-class boundaries, especially for overlapping or ambiguous emotions such as Fear, Surprise, and Disgust. This demonstrates GRACE’s superior ability to extract emotion-discriminative features. Corresponding UAR and WAR scores also indicate performance gains.}
\end{figure}

Unlike global or attention-based matching, our OT formulation explicitly models soft token-to-token correspondences between visual and textual modalities, enabling the network to establish interpretable, emotion-sensitive associations—such as linking “furrowed brows” to specific frame regions. This aligns with our design principle in the Introduction: rather than treating alignment as a black-box operation, we introduce emotion-aware constraints and selective transport weights to guide matching toward semantically meaningful facial dynamics. The substantial performance gap demonstrates that such structured alignment is not only more accurate but also better suited for the subtle, ambiguous nature of dynamic facial expressions.

Collectively, these fine-grained ablation studies validate the necessity and effectiveness of each major component in the GRACE framework.

\begin{figure*}[!t]
\centering
\includegraphics[width=\linewidth, trim=15 54 15 85, clip]{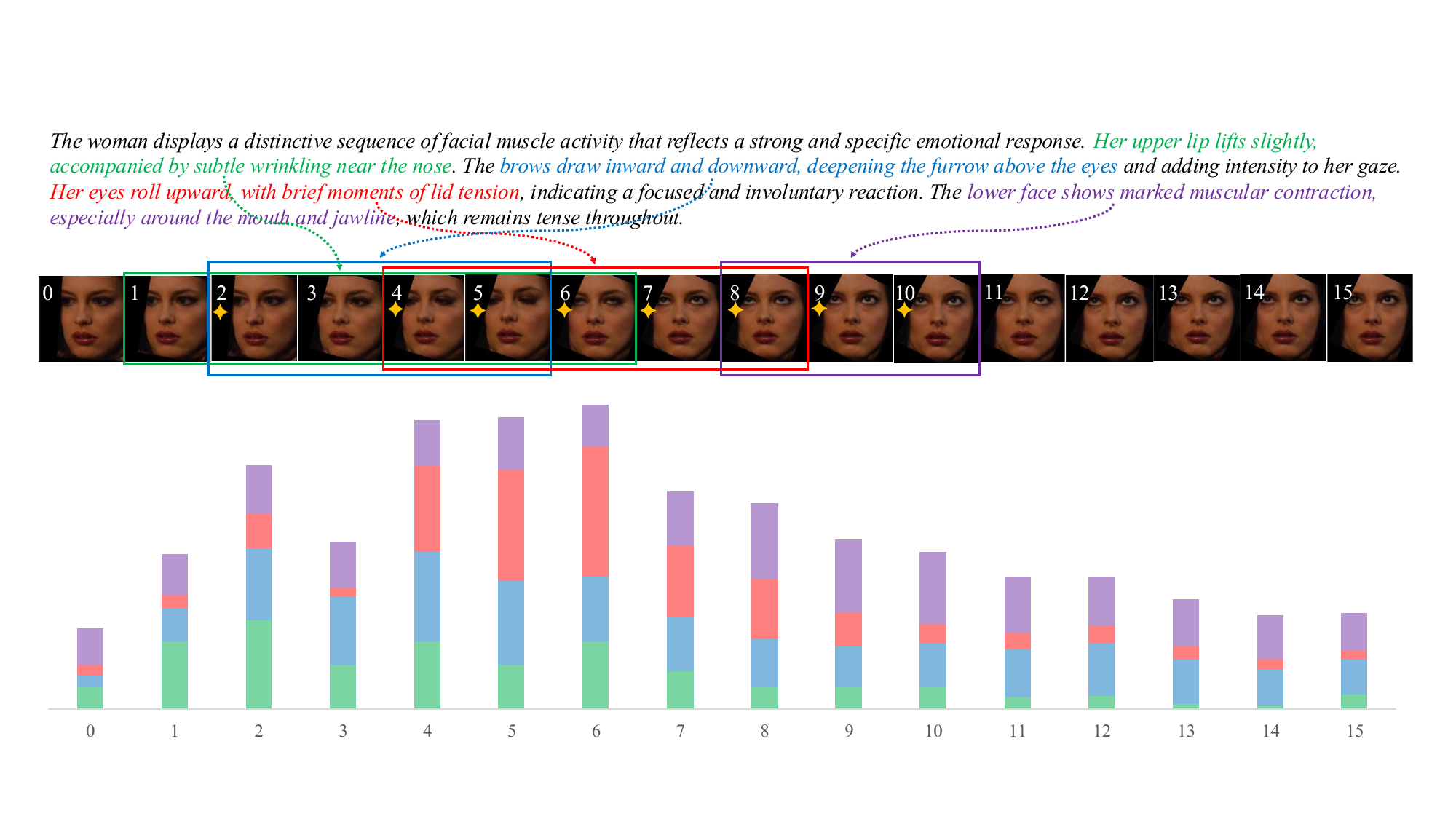}
\caption{Visualization of the GRACE alignment process. Each color-coded phrase in the expression description is aligned with the most relevant video frame via OT transport weights.  Colored boxes indicate frames with the highest alignment to specific phrases. The stacked bar chart shows OT weights across frames, where segment colors match the corresponding phrases. Phrases are derived from BERT tokenizer spans, and token-level OT weights are aggregated to compute phrase-frame alignment scores. The top 8 frames with the highest cumulative scores, marked with \includegraphics[height=2.0ex]{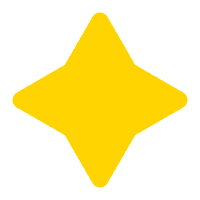} , are selected as key temporal segments for classification.}
\label{fig:visualization_OT}
\end{figure*}

\section{Visualization}

\textbf{Comparative t-SNE Visualization of Learned Representations.}
To further evaluate the quality of learned representations, we visualize the emotion embeddings using t-SNE for both the Baseline method and our GRACE. As shown in Fig.~\ref{fig:tsne_visualization}, the embeddings from GRACE are globally more dispersed across the space, yet each emotion class forms more cohesive and well-separated clusters. In particular, difficult-to-distinguish classes such as \textit{Fear} and \textit{Disgust} exhibit significantly reduced overlap, suggesting improved class separability. In contrast, the Baseline model yields more entangled feature distributions, with extensive inter-class mixing—especially among \textit{Neutral}, \textit{Angry}, and \textit{Surprise}. These observations are aligned with the quantitative metrics, where GRACE achieves 68.42\% UAR and 76.29\% WAR, outperforming the Baseline by 7.38\% and 2.48\%, respectively. This demonstrates that our semantically guided framework effectively enhances emotional discrimination at the feature level, especially under ambiguous or imbalanced conditions.

\textbf{Optimal Transport alignment visualization.}
To further evaluate the interpretability and effectiveness of our OT-based alignment mechanism, we provide a qualitative visualization in Fig.~\ref{fig:visualization_OT}. Although the OT alignment is performed at the token level, we manually group a subset of contiguous tokens into semantically meaningful spans based on their positions in the BERT tokenizer output. For each span, we aggregate the OT transport weights of its constituent tokens over all frames. 

As shown in the figure, our model aligns expression-relevant textual phrases—such as \textit{“upper lip lifts slightly”}, \textit{“brows draw inward and downward”}, and \textit{“eyes roll upward”}—with the most relevant frames in the video. Colored bounding boxes highlight the frames with the highest transport weight for each phrase, illustrating the model’s ability to localize subtle but emotionally salient cues within the temporal sequence.

The stacked bar chart below reflects the full distribution of OT weights across frames for all selected phrases, revealing a coherent soft alignment pattern rather than sparse or isolated mappings. Crucially, the final key clips used for classification are selected by ranking cumulative OT scores across all tokens, thereby aggregating the semantic contributions of multiple expression-relevant spans. This visualization demonstrates how fine-grained linguistic cues contribute collectively to the selection of temporally localized, semantically aligned key segments via OT matching.

\section{Limitations}
\label{sec:limitations}

While the proposed GRACE framework demonstrates strong performance across multiple dynamic facial expression benchmarks, several limitations remain to be addressed in future work.

First, the overall effectiveness of our method is inherently dependent on the quality of both visual and textual feature representations. Although we leverage pretrained models such as VideoMAE and VideoCLIP, the semantic richness and granularity of extracted features—and consequently the alignment accuracy—remain bounded by the representational capacity of these backbone networks. Improving foundational visual and language encoders is thus crucial for further advancement.

Second, our method selects salient features for alignment by retaining a fixed proportion of top-ranked tokens based on saliency scores. This fixed-ratio strategy, while effective in general cases, may inadvertently preserve noisy or non-informative regions, particularly in ambiguous or compound expressions. An adaptive or context-aware selection mechanism could help mitigate this issue.

Third, the proposed token-level alignment relies on fine-tuning a pretrained multimodal model~\cite{xu2021videoclip}, as current general-purpose VL models are not optimized for the granularity and temporal sensitivity required in facial expression alignment. Developing more flexible and robust token-level alignment frameworks for spatiotemporal affective data remains an open challenge.

Lastly, our approach does not explicitly incorporate contextual information beyond the facial region. Given that different individuals may exhibit similar dynamics for distinct emotions—or conversely, exhibit distinct dynamics for the same emotion—expression recognition may benefit from incorporating auxiliary cues such as scene context, speaker identity, or prior affective history. Future work may explore integrating such structured background information to provide stronger semantic priors for emotion understanding.

\section{Conclusion}
\label{sec:conclusion}

In this work, we propose GRACE, a semantically grounded framework that integrates motion-aware visual cues and emotion-anchored textual representations for dynamic facial expression recognition. Unlike prior approaches that rely on coarse labels or global feature matching, GRACE introduces three key innovations: (1) CATE to emphasize emotionally discriminative semantic units, (2) motion-difference-based visual filtering to isolate expressive facial dynamics, and (3) token-level cross-modal alignment via entropy-regularized optimal transport.

Comprehensive experiments on three benchmark datasets—DFEW, FERV39k, and MAFW—demonstrate that GRACE consistently outperforms SOTA methods, particularly on semantically sparse or visually ambiguous categories. Ablation studies further validate the individual and complementary contributions of each proposed component.

By bridging linguistic structure and spatiotemporal expression patterns through fine-grained, interpretable alignment, GRACE sets a new direction for emotion recognition research. Future efforts may build on this foundation by improving alignment generality, incorporating broader contextual cues, and advancing feature extraction backbones tailored for affective understanding.

% \bibliographystyle{IEEEtran}
% \bibliography{bib_paper}

\end{document}